\def\ie{\textit{i.e.}}
\def\eg{\textit{e.g.}}
\def\wrt{\textit{w.r.t. }}
\def\vs{\textit{v.s. }}
\documentclass[runningheads]{llncs}
\usepackage{graphicx}

\usepackage{tikz}
\usepackage{comment} 
\usepackage{amsmath,amssymb} 
\usepackage{color}
\usepackage{color}
\usepackage{hyperref}
\usepackage{wrapfig}
\usepackage{multirow}
\usepackage{booktabs}
\usepackage{array}
\usepackage{bm}
\usepackage{float}

\begin{document}
\pagestyle{headings}
\mainmatter
\def\ECCVSubNumber{1290}  

\title{CycAs: Self-supervised Cycle Association for Learning Re-identifiable  Descriptions} 


\titlerunning{CycAs: Self-supervised Cycle Association for Re-ID}
%
\author{Zhongdao Wang\inst{1} \and
Jingwei Zhang\inst{1} \and
Liang Zheng\inst{2} \and
Yifan Sun\inst{3} \and
Yali Li\inst{1} \and
Shengjin Wang\inst{1}
}
\authorrunning{Z. Wang et al.}
%
\institute{Department of Electronic Engineering, Tsinghua University 
\email{wcd17@mails.tsinghua.edu.cn, \{liyali13,wgsgj\}@tsinghua.edu.cn}\\\and
Australian National University \email{liang.zheng@anu.edu.au} \and MEGVII Technology \email{peter@megvii.com}
}
\maketitle
\begin{abstract}
This paper proposes a self-supervised learning method for the person re-identification (re-ID) problem, where existing unsupervised methods usually rely on pseudo labels, such as those from video tracklets or clustering. 
A potential drawback of using pseudo labels is that errors may accumulate and it is challenging to estimate the number of pseudo IDs. 
We introduce a different unsupervised method that allows us to learn pedestrian embeddings from raw videos, without resorting to pseudo labels. 
The goal is to construct a self-supervised pretext task that matches the person re-ID objective. 
Inspired by the \emph{data association} concept in multi-object tracking, we propose the \textbf{Cyc}le \textbf{As}sociation (\textbf{CycAs}) task: after performing data association between a pair of video frames forward and then backward, a pedestrian instance is supposed to be associated to itself. 
To fulfill this goal, the model must learn a meaningful representation that can well describe correspondences between instances in frame pairs. 
We adapt the discrete association process to a differentiable form, such that end-to-end training becomes feasible. 
Experiments are conducted in two aspects:
We first compare our method with existing unsupervised re-ID methods on seven benchmarks and demonstrate CycAs' superiority. 
Then, to further validate the practical value of CycAs in real-world applications, we perform training on self-collected videos and report promising performance on  standard test sets. 

\end{abstract}

\section{Introduction}
Self-supervised learning is a recent solution to the lack of labeled data in various computer vision areas like optical flow estimation~\cite{flow1,flow2}, disparity/depth estimation~\cite{mde1,mde2,disp1}, pixel/object tracking~\cite{cycle1,cycle2,dut} and universal representation learning~\cite{ss1,ss2,ss3,ss4,ss5,ss6}. As a branch of unsupervised learning, the idea of self-supervised learning is to construct a \emph{pretext task}. It is supposed that free supervision signals of the pretext task can be generated directly from the data, and the challenges lie in the design of the pretext task, so that the learned representation matches the task objective.


Self-supervised / unsupervised learning finds critical significance in the person re-identification (re-ID) area, because of the high annotation cost. 
The goal of the re-ID task is to search for cross-camera bounding boxes containing the same person with a query: the cross-camera requirement increases the burden of data annotation.
Existing unsupervised methods usually rely on pseudo labels that can be obtained from video tracklets~\cite{uga,taudl} or clustering~\cite{pul,buc}. This strategy achieves descent accuracy, but its potential drawback consists of the error accumulation and the challenge in estimating the number of pseudo identities. 





\begin{figure}[t]
    \centering
    \includegraphics[width=\linewidth]{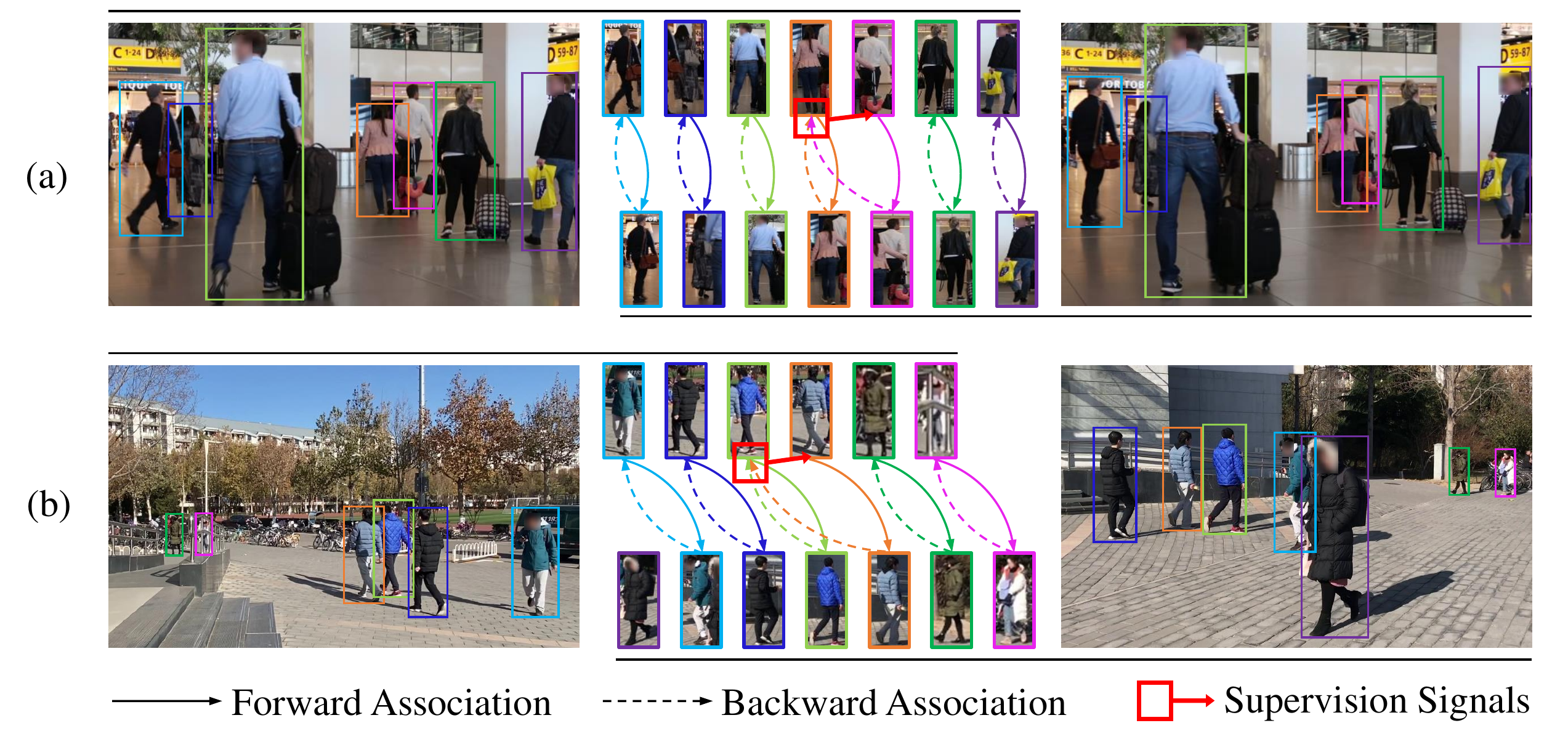}
    
    \caption{Cycle association between (a). Temporal consecutive frames in a single video; (b). Temporal-aligned frames from two cameras that shares an overlapped visual field. In these examples, (a) shows a symmetric case (two frames contain the same group of identities) and (b) shows an asymmetric case.}
    \label{fig:intro}
    
\end{figure}


We are interested in finding a pretext task for re-ID, such that pedestrian descriptors can be learned in a self-supervised way. 
We are motivated by the cycle association between a pair of video frames that contain multiple persons. 
Considering two temporal-consecutive frames from such a video, because of the short time interval between them, they usually share the same group of identities (Figure~\ref{fig:intro} (a)).
With perfect person representations, if we apply data association\footnote{
In Multi-Object Tracking (MOT)~\cite{mot15}, data association means matching observations in a
new frame to a set of tracked trajectories. In our case, we simplify the concept to matching observations between a frame pair.} between the two frames, we can find accurate correspondences between the two sets of identities. 
Further, if we perform forward data association and then backward, an instance is supposed to be associated to itself. 
Based on this motivation, we construct the \textbf{Cy}cle \textbf{As}sociation (CycAs) pretext task: apply data association in a cycle, \ie, forward then backward, and use the inconsistency in the cycle association matrix as supervision signals. 
To maximize the cycle consistency or minimize inconsistency, the model inclines to learn a meaningful representation that can well-describe correspondences between instances~\cite{cycle1,cycle2}.

In the above pretext task, we face a dilemma, \emph{i.e.,} appearance diversity vs. accurate association. 
To learn robust descriptors, we need a pedestrian to exhibit sufficient diversity between two frames, \emph{e.g., } people from frames with long temporal interval. It means that persons from the two frames do not come from the exact same group of identities, creating asymmetry and leading to inaccurate association. On the other hand, if we use consecutive frames from a single video to ensure accurate association, the appearance variation would be small, compromising the descriptor robustness. 
To address this dilemma, we modify the optimization objective of CycAs by a relaxation to allow tolerance to a moderate level of asymmetry. Moreover, we adopt a two-stage training procedure: first train the pretext task using consecutive video frames, and then for fine-tuning add frame pairs from different cameras with overlapping field of view (FOV) (see Figure~\ref{fig:intro} (b)).
This training process allows learning correspondences across cameras and benefits feature robustness to large appearance variation, an essential requirement of re-ID.

Experiments are conducted in two aspects:
First, we compare with existing unsupervised re-ID methods on a wide range of public datasets under the same train / test protocol.
Second, to further validate the practical value of the proposed method in real-world applications, we train CycAs with self-collected videos and conduct cross-domain evaluation on Market-1501~\cite{market} and DukeMTMC-ReID~\cite{duke}. Very promising results are shown compared with some direct transfer baselines. 
Our strengths are summarized below,
\begin{itemize}
    \item [(1)] We propose CycAs, a self-supervised pretext task for person re-ID. Strong features can be learned from associating persons between videos frames.
    \item [(2)] We design a relaxed optimization objective for CycAs, allowing leveraging frames with large appearance variation. It significantly improves the discriminative ability of learned representations.
    \item [(3)]  We showcase the strength of CycAs on public benchmarks. We further validate the practical value of CycAs using self-collected videos as training data.  
\end{itemize}





\section{Related Work}
\noindent \textbf{Unsupervised Person re-ID.} Most existing deep learning based unsupervised person re-ID approaches in literature can be categories into there paradigms:
\begin{itemize}
    \item [(1).] { Domain adaptation methods~\cite{HHL,SPGAN,tjaidl,ecn}. These methods start with a supervised learned model which is pre-trained using the source domain data, and then transfer knowledge from the unlabeled target domain  data.}
    \item [(2).] { Clustering-based methods~\cite{pul,buc,CDS}. These methods usually adopt the iterative clustering-and-training strategy. Unlabeled data are grouped with clustering algorithms and assigned pseudo labels, then the model is fine-tuned with these pseudo labels. Such procedure repeats until convergence.}
    \item [(3).] { Tracklet-based methods~\cite{taudl,utal,uga}. These methods label different trackelts, from a specific camera, as different identities, and train multiple classification tasks for multiple cameras in a parallel manner. Cross-camera matching is usually modeled as metric learning with pseudo labels. }
\end{itemize}

Domain adaptation methods need supervised learned pre-train models so they are not totally unsupervised in essence. Clustering/Tracklet-based methods all rely on pseudo labels. In contrast, the proposed CycAs is unsupervised and does not require pseudo labels.

\noindent \textbf{Self-supervised Learning.} As a form of unsupervised learning, self-supervised learning seeks to learn from unlabeled data by constructing pretext tasks. For instance, image-level pretext tasks such as predicting context~\cite{ss1}, rotation~\cite{ss2} and color~\cite{ss5} are useful for learning universal visual representations. 

Video-level pretext tasks~\cite{ss3,cycle1,cycle2} are recently prevalent due to the large amount of available web videos and the fertile information they  contain. Our work is closely related to a line of video-based self-supervised methods that utilize cycle consistency as free supervision~\cite{cycle1,cycle2,dut}. While these methods usually focus on learning fine-grained correspondences between pixels~\cite{cycle1,cycle2}, thus mainly tackle the tracking problem, ours focus more on learning high-level semantic correspondences and is more adaptive to the re-ID problem. To the best of our knowledge, we are the first to provide a self-supervised solution to learn re-identifiable object descriptions.


\section{Proposed Cycle Association Task}
\subsection{Overview}
\begin{figure}[t]
    \centering
    \includegraphics[width=0.9\linewidth]{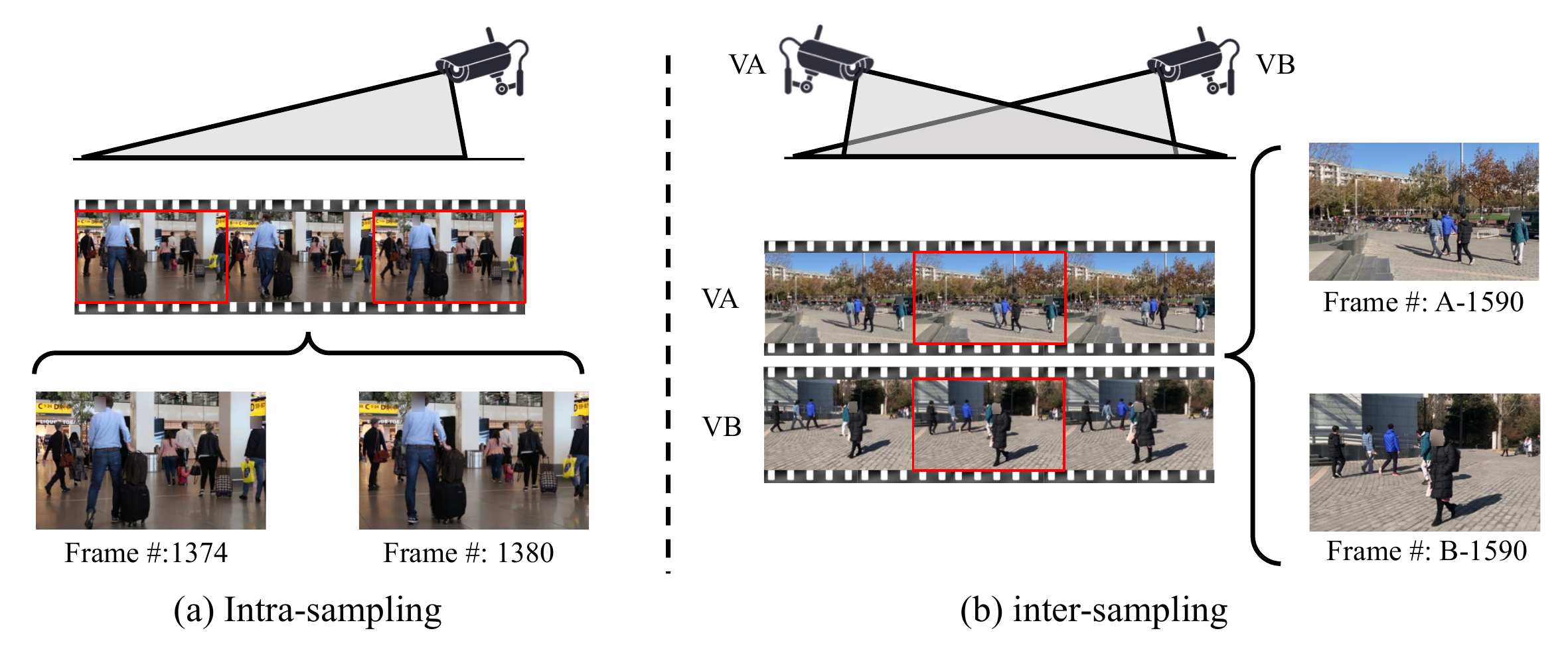}
    \caption{Illustration on the two frame pair sampling methods. (a) Intra-sampling, frame pairs are drawn from a single video within a short temporal interval. (b) Inter-sampling, frame pairs are drawn from cameras capturing an overlapped FOV, at the same time.  }
    \label{fig:sampling}
\end{figure}

 Our goal is to learn a discriminative pedestrian embedding function $\Phi$ by learning correspondences between two sets of person images $\bm{I}_1$ and $\bm{I}_2$. Specifically, $\bm{I}_1$ and $\bm{I}_2$ are detected pedestrian bounding boxes in a pair of frames. In this paper, we design two strategies to sample the frame pairs as follows (Figure~\ref{fig:sampling}).

\begin{itemize}
    \item \textbf{Intra-sampling.} Frame pairs are sampled from the same video within a short temporal interval, \ie, 2 seconds.
    \item \textbf{Inter-sampling.} Each frame pair is sampled at the same timestamp from two different cameras that capture an overlapped FOV.
\end{itemize}

\begin{figure}[t]
    \centering
    \includegraphics[width=\linewidth]{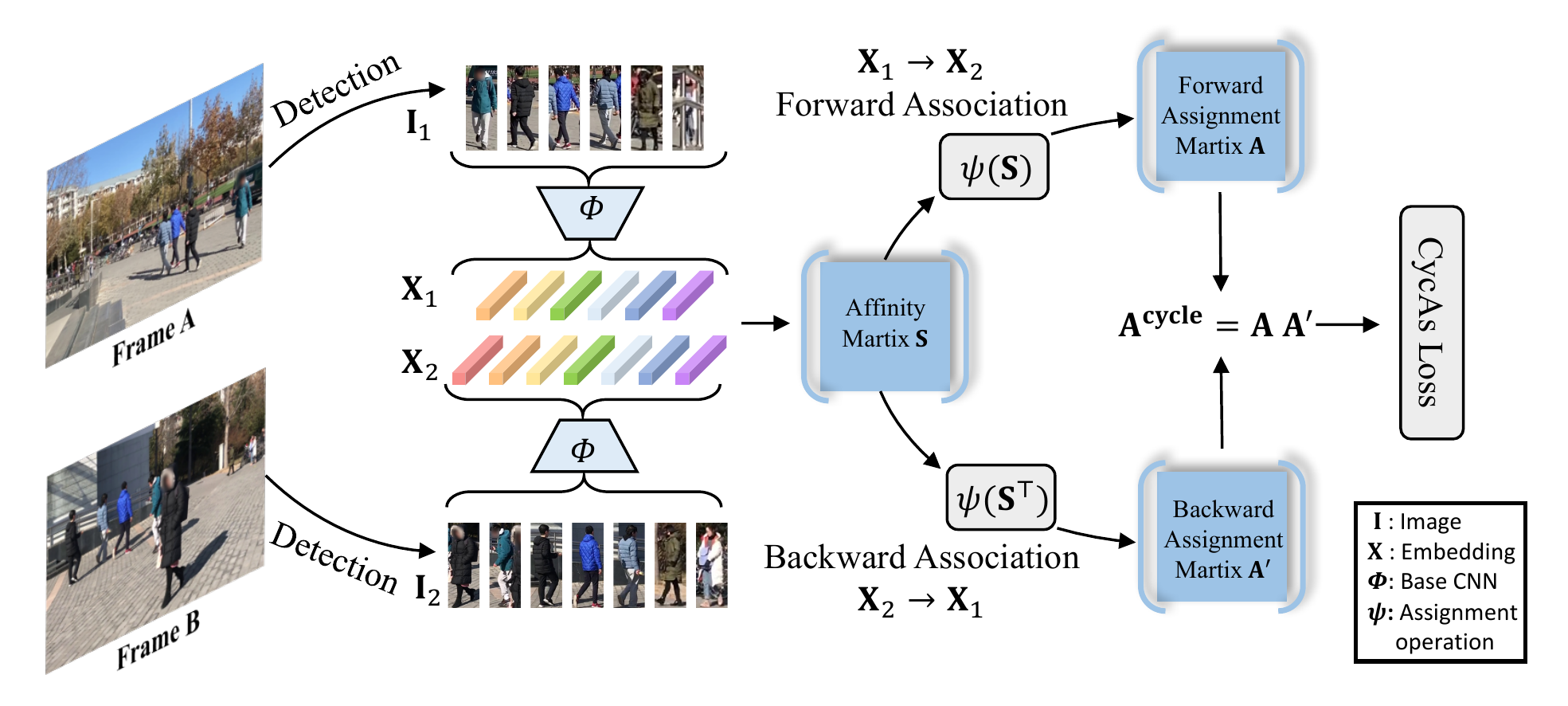}
    \caption{An overview of the proposed cycle association task. First, two sets of detected pedestrians are mapped to embeddings via the base CNN $\Phi$. Pairwise affinity matrix $\bm{S}$ is computed from the two sets of embeedings, then forward/backward assignment matrix $\bm{A}$ and $\bm{A'}$ if computed from $\bm{S}$. Finally the inconsistency in the cycle association matrix $\bm{A}^\texttt{cycle} =\bm{AA'} $ is used as supervision signals.}
    \label{fig:overview}
\end{figure}

In both strategies, with proper selection of temporal interval or deployment of cameras, which should not be too difficult to control, a reasonable identity overlap between $\bm{I}_1$ and $\bm{I}_2$ can be guaranteed. We define $\tau = \frac{\# \; overlapped \; IDs}{max\{ \vert \bm{I}_1 \vert, \vert \bm{I}_2 \vert \}}$ as the \emph{symmetry} between $\bm{I}_1$ and $\bm{I}_2$, and its impact on our system is investigated in Section~\ref{sec:exp:simu}. For ease of illustration, let us begin with the absolute symmetric case  $\tau=1$, \ie, for any instance in $\bm{I}_1$ there is a correspondence in $\bm{I}_2$,  and vice versa. Then we introduce how we deal with asymmetry in Section~\ref{sec:method:asymmetry}. The training procedure is presented in Figure~\ref{fig:overview}.

\subsection{Association Between Symmetric Pairs}
Consider all the images in $\bm{I}_1 \cup \bm{I}_2$ forming a minibatch. Suppose the size of the two sets $\vert \bm{I}_1 \vert = \vert \bm{I}_2 \vert = K$ ($\tau = 1$, \emph{i.e.,} absolute symmetry). The bounding boxes are mapped to the embedding space by $\Phi$, such that $\bm{X}_1 = \Phi (\bm{I}_1)$ and $\bm{X}_2 = \Phi (\bm{I}_2)$, where
$\bm{X}_1 = [\bm{x}_1^1, \bm{x}_1^2,...,\bm{x}_1^{K}] \in \mathbb{R}^{D\times K}$ and 
$\bm{X}_2 = [\bm{x}_2^1, \bm{x}_2^2,...,\bm{x}_2^{K}] \in \mathbb{R}^{D\times K}$ are embedding matrices composed of $K$ embedding vectors of dimension $D$. 
All the embedding vectors are $\ell_2$-normalized. To capture similarity between instances, we compute an affinity matrix between all instances in $\bm{X}_1 $ and $ \bm{X}_2$ by calculating the pairwise cosine similarities, 
\begin{equation}
\label{eq:sim}
     \bm{S} = \bm{X}_1^\top \bm{X}_2 \in \mathbb{R}^{K\times K}.
\end{equation}

We take $\bm{S}$ as input to perform data association, which aims to predict correspondences in $\bm{X}_2$ for each instance in $\bm{X}_1$. Formally, the goal is to obtain an assignment matrix,
\begin{equation}
    \bm{A} = \bm{\psi} (\bm{S}) \in \{0,1\}^{K\times K},
\end{equation}
 where $1$ indicates correspondence. In MOT, it is usually modeled as a linear assignment problem, and the solution $\bm{\psi}$ can be found by the Hungarian algorithm (examples can be found in many MOT algorithms~\cite{deepsort,poi,jde}). 

Suppose the embedding function $\Phi$ is perfect, \ie, the cosine similarity between vectors of the same identity equals $1$, while the cosine similarity between vectors from different identities equals $-1$. The Hungarian algorithm can output the optimal assignment $\bm{A}^{*}=\frac{\bm{S} + 1}{2}$ for the forward association process $\bm{X}_1 \rightarrow \bm{X}_2$. The backward association process $\bm{X}_2 \rightarrow \bm{X}_1$ is similar, and the optimal assignment matrix $\bm{A}'^{*} =  \frac{\bm{S}^\top + 1}{2}$. 
A Cycle Association pass is then defined as a forward association pass plus a backward association pass,
\begin{equation}
\label{eq:cycle}
    \bm{A}^\texttt{cycle} = \bm{A} \bm{A}'.
\end{equation}

Intuitively, if $\bm{A}=\bm{A}^{*}$ and $\bm{A}'=\bm{A}'^{*}$, an instance will be associated to itself. In other words, the cycle association matrix underpinning perfect association
$\bm{A}^\texttt{cycle}$ should equal the identity matrix $\bm{I}$. 
Accordingly, the difference between $\bm{A}^\texttt{cycle}$ and $\bm{I}$ can be used as signals to implicitly supervise the model to learn correspondences between $\bm{X}_1 $ and $\bm{X}_2$. 

The whole process needs to be differentiable for end-to-end training. However, the assignment operation $\psi$ (Hungarian algorithm) is not differentiable. This motivates us to design a differentiable $\psi$. We notice that if the one-to-one correspondence constraint is removed, $\psi$ can be approximated by the row-wise \texttt{argmax} function. Considering \texttt{argmax} is not differentiable either, we further soften this operation by the row-wise softmax function. Now, the assignment matrix is computed as,
\begin{equation}
\label{eq:softmax}
    \bm{A}_{i,j} = \psi_{i,j} (\bm{S}) = \frac{e^{T\bm{S}_{i,j}}}{\sum_{j'}^K  e^{T{\bm{S}_{i, j'}}}},
\end{equation}
where $\bm{A}_{i,j}$ is the element of $\bm{A}$ in the $i$-th row and the $j$-th colomn, and $T$ is the temperature of the softmax operation. The backward association pass has a different temperature $T'$. $T$ and $T'$ are designed to be adaptive to the size of $\bm{A}$ and $\bm{A}'$, and more details will be described in Section~\ref{sec:method:temp}.

Combing Eq.~\ref{eq:sim}, Eq.~\ref{eq:softmax} and Eq.~\ref{eq:cycle}, a cycle association matrix $\bm{A}^\texttt{cycle}$ can be computed with all operations therein being differentiable. Finally, the loss function is defined as the mean $\ell_1$ error between $\bm{A}^\texttt{cycle}$ and $\bm{I}$,
\begin{equation}
\label{eq:symmetricloss}
    \mathcal{L}_\texttt{symmetric} = \frac{1}{K^2} \Vert \bm{A}^\texttt{cycle} - \bm{I} \Vert_1.
\end{equation}

\begin{figure}[t]
    \centering
    \includegraphics[width=\linewidth]{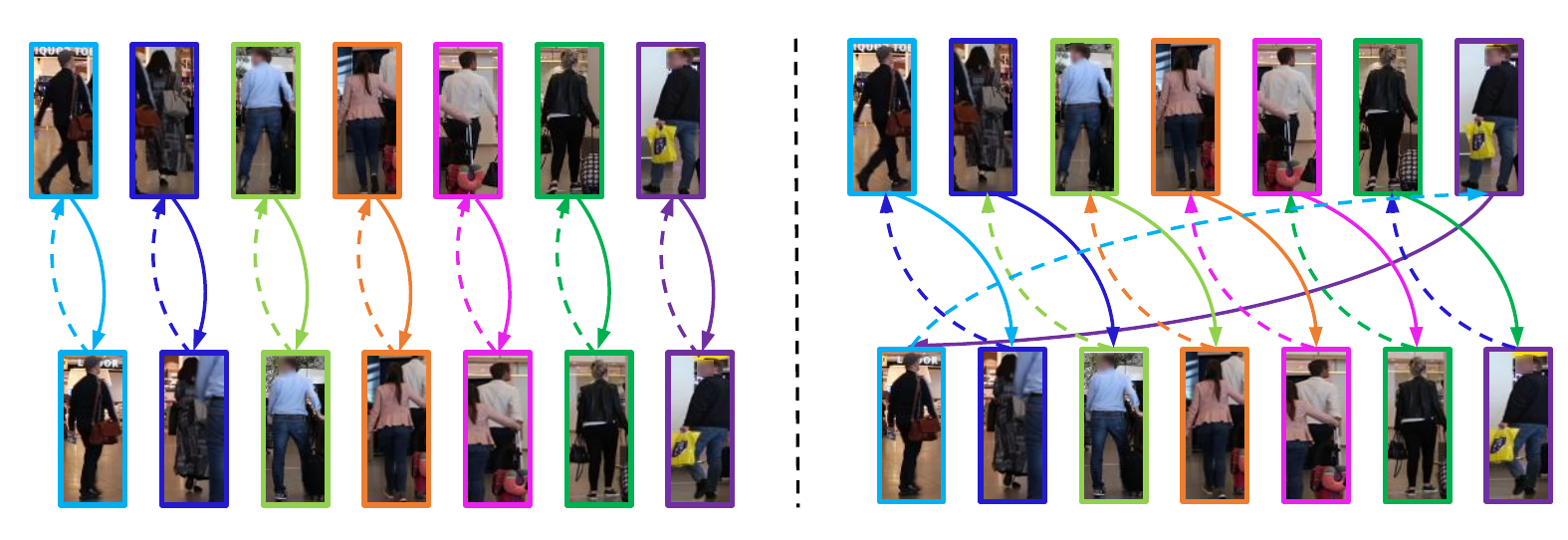}
    \caption{A non-trivial solution (\textbf{Left}) \vs a trivial solution (\textbf{Right}) of Eq. \ref{eq:symmetricloss}. Solid and dotted lines means forward and backward correspondences, respectively. We argue the trivial solutions are very unlikely to be learned.}
    \label{fig:trivial}
\end{figure}

\textbf{Discussion.} Theoretically, {cycle consistency} is a \emph{necessary} but not \emph{sufficient} requirement for discriminative embeddings. 
In Figure~\ref{fig:trivial} we present a trivial solution: what the embedding model has learned is matching an identity to the next identity, and the last identity is matched to the first. Such a trivial solution requires the model to learn strong correlations between random identity pairs, which share very limited, if any, similar visual patterns. Therefore, we reasonably argue that by optimizing the cycle association loss, it is very unlikely for the model to converge to such trivial solutions and that it is much easier to converge to non-trivial solutions, \ie, the discriminative embeddings. Actually, this argument is proved by our experiment: all the converged solutions are disciminative embeddings, and trivial solutions like Figure~\ref{fig:trivial} never emerge. 

\subsection{Relaxation on asymmetric pairs}
\label{sec:method:asymmetry}
In practice, asymmetric always arises along with large appearance diversity.
 The main reasons are multi-folds: First, pedestrians enter and leave the visual field of the camera; second, real-world videos usually contain high-velocity motions, severe occlusions, and low-quality persons, so the detector may fail sometimes; for inter-sampled data, it's also impossible to ensure FOVs overlap exactly. For better leveraging such appearance-diverse data, we make the following efforts to reduce the negative impact of the asymmetry.

First, consider descriptor matrices $\bm{X}_1$ and $\bm{X}_2$ with the number of descriptors being $K_1$ and $K_2$, respectively. According to Eq. \ref{eq:cycle}, the resulting $\bm{A}^\texttt{cycle}$ is of size $K_1 \times K_1$. If $K_1 > K_2$, there will exist at least $K_1 - K_2$ instances that cannot be associated back to themselves. This will introduce ambiguity. Therefore, we swap $\bm{X}_1$ and $\bm{X}_2$ in such cases to ensure $K_1 \le K_2$ always holds.

Second, the learning objective is modified. In the asymmetric scenario, the loss function $\mathcal{L}_\texttt{symmetric}$ is sub-optimal, because some instances may lose correspondences in cycle association, and thus the corresponding diagonal elements in $\bm{A}^\texttt{cycle}$ are not supposed to equal $1$. Simply changing the supervision of these lost instances from $1$ to $0$ is not feasible, because there are no annotations and we do not know which instances are lost. 
To address this, our solution is to relax the learning objective. More specifically,
we expect a diagonal element $\bm{A}^{\texttt{cycle}}_{i,i}$ to be greater than all the other elements along the same row and column, by a given margin $m$. The loss function is formulated as,
\begin{equation}
\label{eq:asymmetricloss}
\scriptsize{
    \mathcal{L}_\texttt{asymmetric} = \frac{1}{K_1} 
    \sum_{i=1}^{K_1} 
    \left [ 
    \left(\max_{j\ne i} \bm{A}^{\texttt{cycle}}_{i,j} - 
    \bm{A}^{\texttt{cycle}}_{i,i} + m\right)_+ +
    \left(
    \max_{k\ne i} \bm{A}^{\texttt{cycle}}_{k,i} - \bm{A}^{\texttt{cycle}}_{i,i} + m \right)_+
    \right ],
    }
\end{equation}
which has a similar form as the triplet loss~\cite{triplet}.  The margin $m$ is a hyper-parameter ranging in $(0,1)$ with smaller values indicating softer constrains. We set $m=0.5$ in all the experiment if not specified. 

We will show through experiment that the relaxation of the loss function benefits learning in both the asymmetric and symmetric cases. 
In the experiment, we use $\mathcal{L}_\texttt{asymmetric}$ by default unless specified.

\subsection{Adapt softmax temperature to varying sizes}
\label{sec:method:temp}
Cosider two vectors with different sizes, $\bm{v}=(1,0.5)^\top$ and $\bm{u}=(1,0.5, 0.5)^\top$. Let $\bm{\sigma}$ be the softmax operation, then $\bm{\sigma(v}) = (0.62,0.38)^\top$ and $\bm{\sigma(u}) = (0.45, 0.27, 0.27)^\top$. 
The \emph{Soft-Max} operation, as we observe, has different levels of softening ability on inputs with different sizes.
The Max value in a longer vector is less highlighted, or maxed, and vice versa. To alleviate this problem and stabelize the training, we let the softmax temperature be adaptive to the varying input size, so that for input vectors of different sizes, the max values in them are equally highlighted.

To fulfill this goal, we let the temprature $T = \frac{1}{\epsilon} \log \left[ \frac{\delta (K-1) + 1}{1-\delta} \right]$, where $\epsilon$ and $\delta$ are two hyper-parameters ranging from 0 to 1. In fact, the only hyper-parameter that matters is $\epsilon$, and $\delta$ can be simply set to $0.5$. This leads to the final form $T = \frac{1}{\epsilon} \log (K+1)$. Detailed derivation and discussion can be found in the supplementary material.

\subsection{Two-stage Training}
\label{sec:method:training}
The two types of sampling strategies have their respective advantages and drawbacks. To get the best of both worlds, we design a two-stage training procedure that initializes from a model pretrained on ImageNet~\cite{imagenet}. 

In Stage I, we train the model with intra-sampled data only, and the temporal interval for sampling is set to rather small, \eg, 2 seconds, so that the appearance variation between two person sets is small, or the symmetric $\tau$ is high ($> 0.9$).

After convergence of Stage I, we start Stage II. In this stage, we train the model using both the inter-sampled data and intra-sampled data in a multi-task manner with $1:1$ loss weights. The inter-sampled data have much higher appearance variations but a lower $\tau$ (around $0.6$).

\textbf{Discussions.} This training strategy is carefully designed so as to converge well.
Directly starting from Stage II converges with a slower speed \wrt our progressive training strategy, while starting from inter-sampled data only fails in converging. We will quantitatively demonstrate the effectiveness of this training strategy in Section~\ref{sec:exp:ablation}. 

\begin{figure}[t]
    \centering
    \includegraphics[width=0.8\linewidth]{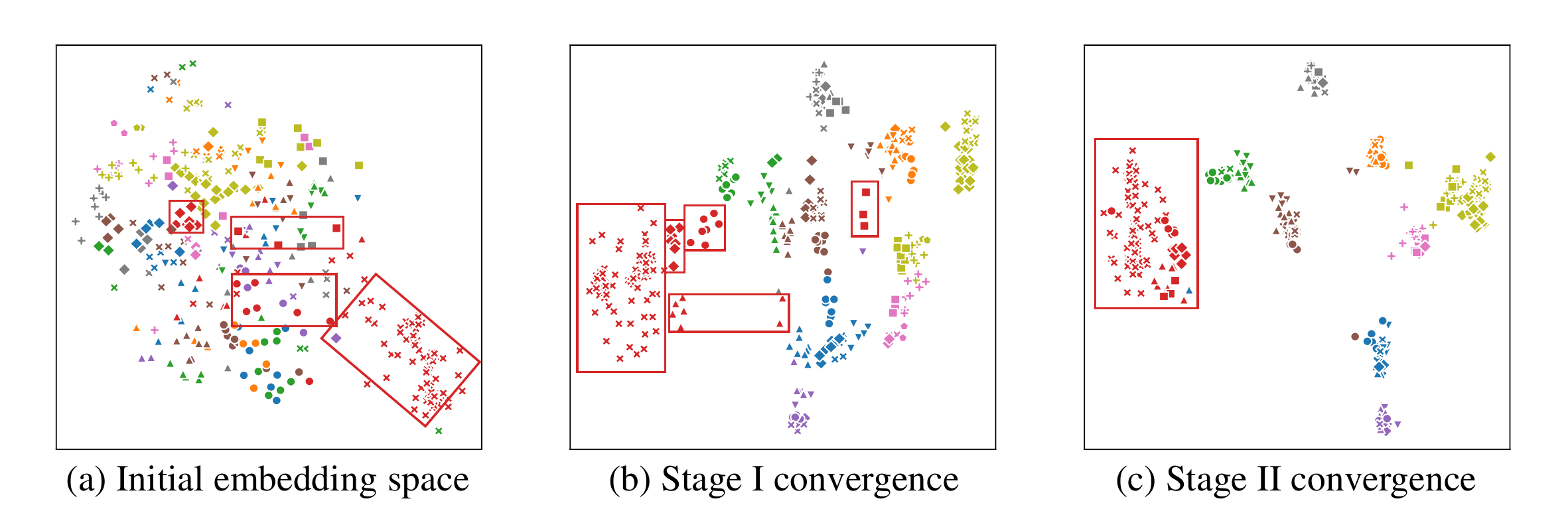}
    \caption{Illustration on the evolution of the embedding space from (a) Initial model to (b) Stage I convergence then to (c) Stage II convergence.  Different colors indicate different identities. Different markers indicate different cameras. Visualized via Barnes-Hut t-SNE~\cite{bhtsne}.}
    \label{fig:tsne}
\end{figure}

Alternatively, we give an intuitive illustration by visualizations on the embedding space at each training state, shown in Figure~\ref{fig:tsne}. In the initial embedding space before training, , most embeddings from different identities are not separable Figure~\ref{fig:tsne}~(a). In Stage I training, the model learns to find correspondence between persons within the same camera, so in the resulting embedding space embeddings of the same identity from the same cameras are grouped together, while embeddings of the same identity from different cameras are still separate (see red rectangles in Figure~\ref{fig:tsne}~(b)). Finally, in Stage II training, the model learns to associate across different cameras, in which case the appearance variations are large. Therefore, the resulting embedding can handle large appearance variation, thus is camera-invariant(Figure~\ref{fig:tsne}~(c)). To summarize, training Stage I functions as a ``warm-up" process for Stage II, while in Stage II the model learns meaningful camera-invariant representation for the re-ID task. 

\section{Experiment}
The experiments are organized as follows.
First, we adopt standard train / test protocols on seven video- / image-based person re-ID datasets and compare our method with existing unsupervised methods (Section~\ref{sec:exp:simu}). 
Then we  investigate the impact of different components and hyper-parameters (Section~\ref{sec:exp:ablation}). 
Finally, to be more practical, we perform experiment using self-collected pedestrian videos as training data and compare with direct transfer (supervised) models (Section~\ref{sec:exp:prac}). 
We use ResNet-50~\cite{resnet} as the backbone network in all experiments.

\subsection{Experiment with Standard Datasets.}
\label{sec:exp:simu}
\textbf{Setup.} We test the proposed CycAs on both video-based (MARS~\cite{mars}, iLIDS~\cite{ilids}, PRID2011~\cite{prid}) and image-based (Market-1501~\cite{market}, CUHK03~\cite{cuhk03}, DukeMTMC-ReID~\cite{duke}, MSMT17~\cite{msmt}) person re-id datasets. All the datasets provide camera annotations and the video-based datasets additionally provide tracklets. Following existing practice \cite{taudl,utal,uga}, for image-based datasets, we assume all images per ID per camera are drawn from a single tracklet.
Consider a mini-batch with batch size $B$, to mimic the intra- / inter-sampling, we first randomly sample $B/2$ identities.
For intra-sampling, a tracklet is sampled for each of these identities; then, two bounding boxes are sampled within each tracklet.
For inter-sampling, two tracklets from different cameras are sampled for each of the $B/2$ identities; then, one image is is sampled from each tracklet. 
Results on image- / video-based re-ID datasets are shown in Table~\ref{tab:exp:image} and Table~\ref{tab:exp:video}, respectively.

   \begin{figure}[t]
     \centering
     \includegraphics[width=\linewidth]{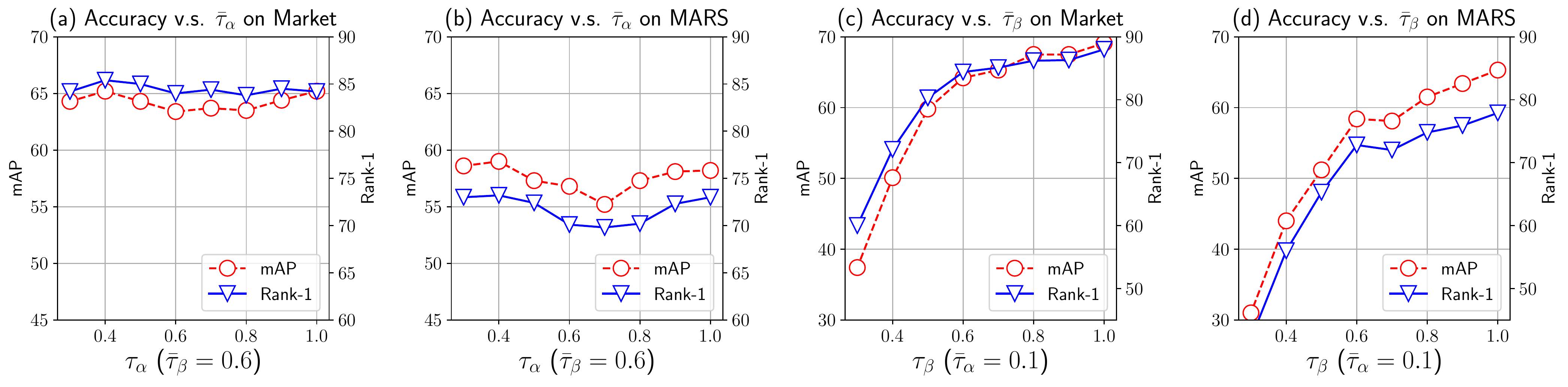}
     \caption{Robustness against different levels of (a-b) intra-sampled data symmetry $\tau_\alpha$ and (c-d) inter-sampled data symmetry $\tau_\beta$. For evaluating $\tau_\alpha$, in each mini-batch, we fix $\tau_\beta$ and draw a random $\tau_\alpha$ from $\mathcal{N}(\bar{\tau}_\alpha, 0.01)$, and plot the performance curve \wrt different mean $\bar{\tau}_\alpha$. The impact of $\tau_\beta$ is evaluated in a similar way. }
   \label{fig:exp:symmetry}
 \end{figure}

\textbf{Performance upper bound analysis.} According to above sampling strategy, the data are absolutely symmetric, \ie, $\tau = 1$. The performance under this setting can be seen as an upper bound of the proposed method, denoted as CycAs$^\texttt{sym}$. 
For comparison, we implement a supervised baseline (IDE~\cite{market}). We observe that the performance of CycAs$^\texttt{sym}$ is consistently competitive on all the datasets. Compared with IDE, the rank-1 accuracy of CycAs$^\texttt{sym}$ is lower by only $1.1\%$, $0.3\%$ and $1.6\%$ on Market-1501, DukeMTMC-ReID and MSMT17, respectively. 
On CUHK03, CycAs$^\texttt{sym}$ even surpasses IDE by $+2.2\%$. This result partially prove the good alignment between the CycAs task and the objective of re-ID. We also list state-of-the-art supervised methods~\cite{pcb,GLTR} for comparison and observe that the performance gap between CycAs$^\texttt{sym}$ and these methods is not too large.
These results suggest the potential of CycAs in the re-ID task.
 

 \begin{table}[tb]

\centering
\scriptsize{
    
    \begin{tabular}{l||c || c ||p{0.75cm}<{\centering} |p{0.75cm}<{\centering} ||  p{0.75cm}<{\centering} |p{0.75cm}<{\centering} ||  p{0.75cm}<{\centering} |p{0.75cm}<{\centering} ||  p{0.75cm}<{\centering} |p{0.75cm}<{\centering}   }
    \hline
           \multirow{2}{*}{Method} & \multirow{2}{*}{Category} & \multirow{2}{*}{Require} &   \multicolumn{2}{c||}{Market~\cite{market}}  & \multicolumn{2}{c||}{Duke~\cite{duke}} & \multicolumn{2}{c||}{CUHK03~\cite{cuhk03}} & \multicolumn{2}{c}{MSMT17~\cite{msmt}} \\
           \cline{4-11}
           &&& R1 & mAP & R1 & mAP & R1 & mAP & R1 & mAP \\
           \hline
           \hline
         SPGAN~\cite{SPGAN} & UDA & Pretrain &   51.5 & 22.8 & 41.1 & 22.3 &- & - & - & - \\
         SPGAN+LMP~\cite{SPGAN} & UDA & Pretrain &   57.7 & 26.7 &  46.4 & 26.2 &- & - & - & - \\
         TJ-AIDL~\cite{tjaidl} & UDA &  Pretrain & 58.2 & 26.5 & 44.3 & 23.0  &- & - & - & - \\
         HHL~\cite{HHL}& UDA & Pretrain & 62.2 & 31.4 & 46.9 & 27.2 & - & - & - & - \\
         ECN~\cite{ecn} & UDA & Pretrain & 75.1 & 43.0 & 63.3 & 40.4 & - & - &  30.2 & 10.2 \\
         \hline
         PUL~\cite{pul}& Clustering & Pretrain & 44.7 & 20.1 &30.4 &16.4  & - & - & - & - \\
         CAMEL~\cite{camel}& Clustering & Pretrain & 54.5 & 26.3& - & - & 39.4 &- &- & -\\
         BUC~\cite{buc}& Clustering & None & 66.2 & 38.3 & 47.4 & 27.5 & - & - & - & - \\
         CDS~\cite{CDS}& Clustering & Pretrain & 71.6 & 39.9 &67.2 & 42.7 & - & - & - & - \\
         \hline
         TAUDL~\cite{taudl} & Tracklet & MOT &  63.7 & 41.2 & 61.7 & 43.5 & 44.7 & 31.2 & 28.4 & 12.5 \\
         UTAL~\cite{utal} & Tracklet & MOT & 69.2 & 46.2 & 62.3 & 44.6 &  \textbf{56.3} & \textbf{42.3} &  31.4 & 13.1 \\
         UGA~\cite{uga} & Tracklet & MOT & \textbf{87.2} & \textbf{70.3} &  \emph{75.0} & \emph{53.3} &  - & - &  \emph{49.5} &\emph{21.7} \\
         \hline
                 \textbf{CycAs$^\texttt{asy}$}& Self-Sup & None & \emph{84.8} & \emph{64.8} & \textbf{77.9} & \textbf{60.1} &  \emph{47.4} & \emph{41.0} & \textbf{50.1} &\textbf{26.7} \\

         \textbf{CycAs$^\texttt{sym}$}& - & None & 88.1 & 71.8 &  79.7 & 62.7&56.4 &  49.6 & 61.8 &  36.2 \\
         \hline
         IDE & Supervised & Label & 89.2 & 73.9 & 80.0 & 63.1 & 54.2 & 47.2 & 60.2 & 33.4 \\
         PCB+RPP~\cite{pcb}& {Supervised} & Label & 93.8 & 81.6 & 83.3 & 69.2 &63.7 &   57.5 & - &  - \\
         \hline

    \end{tabular}
    \caption{Comparison  with state-of-the-art methods on image-based re-ID datasets. Note all the methods starts from a ImageNet pretrained model. The requirement \emph{Pretain} and \emph{None} refers to whether pretraining on labeled re-ID datasets is needed. CycAs$^\texttt{asy}$ is our method, and CycAs$^\texttt{sym}$ refers to an upper bound of our method.}
    \label{tab:exp:image}}



    \centering
    \scriptsize{
    
    \begin{tabular}{l||c || c ||p{1.2cm}<{\centering} |p{1.2cm}<{\centering} ||  p{1.2cm}<{\centering}  ||  p{1.2cm}<{\centering}    }
    \hline
           \multirow{2}{*}{Method} & \multirow{2}{*}{Category} & \multirow{2}{*}{Require} &   \multicolumn{2}{c||}{MARS~\cite{mars}}  & {PRID~\cite{prid}} & {iLIDS~\cite{cuhk03}}  \\
           \cline{4-7}
           &&& R1 & mAP & R1  & R1  \\
           \hline
           \hline
        
         DGM+IDE~\cite{DGM} & Clustering & Pretrain & 36.8 & 21.3 & 56.4 & 36.2 \\
         RACE~\cite{RACE} & Clustering & MOT & 43.2 &  24.5 & 50.6 & 19.3 \\
         BUC~\cite{buc} & Clustering & None & 61.1 &  38.0 & - & - \\
         \hline
         SMP~\cite{SMP} & Tracklet & MOT &   23.9 &  10.5 & 80.9 & 41.7 \\
         TAUDL~\cite{taudl}& Tracklet & MOT & 43.8 & 29.1 & 49.4 & 26.7 \\
         UTAL~\cite{utal} & Tracklet & MOT & 49.9 & 35.2 & 54.7 & 35.1 \\
         UGA~\cite{uga} & Tracklet & MOT &  58.1 &  39.3 & 80.9 & 57.3 \\
        
         \hline
         \textbf{CycAs$^\texttt{asy}$}& Self-Sup & None & \textbf{72.8}  & \textbf{58.4}&  \textbf{86.5} &\textbf{73.3} \\
          \textbf{CycAs$^\texttt{sym}$}& - & None & 79.2 & 67.5 & 85.4 & 77.3  \\
         \hline
         IDE & {Supervised} & Label & 81.7 & 67.8 & 90.5 & 78.4  \\
         GLTR~\cite{GLTR}& {Supervised} & Label & 87.0 & 78.5 & 95.5 & 86.0  \\
         \hline

    \end{tabular}
    \caption{Comparison with state-of-the-art methods on video-based re-ID datasets.}
    \label{tab:exp:video}}

\end{table}

 \textbf{Robustness against different levels of data symmetry $\tau$.} 
 To investigate the robustness of CycAs against different levels of symmetry, we introduce asymmetry to the sampled data and observe how the ReID accuracy changes. 
 Specifically, we control intra-sampling symmmetry $\tau_\alpha$ and inter-sampling asymmetry $\tau_\beta$ by replacing a portion of images in $\bm{I}_2$ with randomly sampled images from irrelevant identities.
 To evaluate the impact of $\tau_\alpha$, in each mini-batch, we fix $\tau_\beta$ and draw $\tau_\alpha$ from a gaussian distribution $\mathcal{N}(\bar{\tau}_\alpha, 0.01)$, truncate the value in range $(0,1)$, and plot the model performance against the mean $\bar{\tau}_\alpha$. The impact of $\tau_\beta$ is evaluated in a similar way. We report results  in Fig. \ref{fig:exp:symmetry}.

Two observations can be made from the curves. First, with a moderate fixed value of $\tau_\beta$, \ie, $0.6$ in our case, the model accuracy is robust to a wide range of $\bar{\tau}_\alpha$. For example, in  Fig.~\ref{fig:exp:symmetry} (a), the rank-1 accuracy is both $84.2\%$ when $\bar{\tau}_\alpha$ is set to $1$ and $0.3$, respectively. 
Second, we observe that the accuracy improves when $\bar{\tau}_\beta$ becomes larger. 
The main reason is explained in Section~\ref{sec:method:training} and Figrure~\ref{fig:tsne}: Training Stage I (Training with intra-sampled data) only functions as a ``warm-up" process , to provide a meaningful initial point for Stage II.  The knowledge learned from intra-sampled data contributes less on the overall performance. Therefore the final accuracy is less sensitive to $\bar{\tau}_\alpha$. In contrast, learning from inter-sampled data aligns with the objective of re-ID task, therefore the final accuracy is more sensitive to $\bar{\tau}_\beta$.

\textbf{Remarks.} Note that in Fig.~\ref{fig:exp:symmetry}~(c-d), the curves drop very slowly when $\tau_\beta =1$ decreases from $1$ to $0.6$. 
This suggests that CycAs has a good ability to handle data with reasonably asymmetry. 
Such a property is valuable, because in practice we can control $\bar{\tau}_\beta$ in a reasonable range (say from $0.6$ to $0.9$), by carefully placing the cameras. Comparing with manually annotating data, this requires less effort.

\textbf{Comparison with the state of the art.} For fair comparisons, we train CycAs under a practically reasonable asymmetric assumption. We fix $\tau_\alpha=0.9$ and $\tau_\beta=0.6$, and compare the results (denoted as CycAs$^\texttt{asy}$) with existing unsupervised re-ID approaches. Three categories of existing methods are compared, \ie, unsupervised domain adaptation (UDA) \cite{SPGAN,tjaidl,HHL,ecn}, 
clustering-based methods~\cite{pul,camel,buc,CDS,DGM,RACE}, and tracklet-based methods~\cite{taudl,utal,uga}.
Beside re-ID accuracy, we also compare another dimension, \ie, ease of use, by listing the requirements of each method in Table~\ref{tab:exp:image} and Table~\ref{tab:exp:video}. 

Under image-based unsupervised learning, CycAs$^\texttt{asy}$ achieves state-of-the-art accuracy on two larger datasets, \ie, DukeMTMC and MSMT17. The mAP improvement over the second best method~\cite{uga} is $+6.8\%$ and $+5.0\%$ on DukeMTMC and MSMT17, respectively. 
On Market-1501 and CUHK03,  CycAs$^\texttt{asy}$ is very competitive to the best performing methods~\cite{uga}.

Under video-based unsupervised learning, CycAs$^\texttt{asy}$ achieves state-of-the-art results on three datasets. The rank-1 accuracy improvement over the second best method is $+14.7\%$, $+5.6\%$ and $+16.0\%$ on MARS, PRID and iLIDS, respectively. 

Comparing with other unsupervised strategies, CycAs requires less external supervision. For example, UDA methods use a labeld source re-ID dataset, and most clustering-based methods need a pre-trained model for initialization, which also uses external labeled re-ID datasets.
The tracklet-based methods do not require re-ID labels, but require a good tracker to provide good supervision signals.Training such a good tracker also requires external pedestrian labels. Note that ImageNet pretraining is needed by all the methods.
CycAs learns person representations directly from videos and does not require any external annotation. Its requiring less supervision and competitive accuracy making it potentially a more practical solution for unsupervised re-ID.

 \subsection{Ablation Study}
 \label{sec:exp:ablation}
 
 \begin{table}[t]
 \begin{minipage}[t]{1.0\linewidth}
 \begin{minipage}[t]{0.5\linewidth}
  \centering
     \makeatletter\def\@captype{table}\makeatother\caption{Comparison between two losses under different data symmetry settings on Market-1501. We see the asymmetric loss always outperforms the symmetric loss no matter with asymmetric or symmetric data.}
     
      \begin{tabular}{l | l| c  | c }
    \toprule
    Data  & Loss & Rank-1 & mAP \\
    \hline
    $ symmetric $ & &  &\\
    $ \tau_\alpha, \tau_\beta=$ & $\mathcal{L}_\texttt{symmetric}$ & 78.9 & 59.1\\
    $(1.0,1.0)$& $\mathcal{L}_\texttt{asymmetric}$ & \textbf{88.1}& \textbf{71.8}\\
    \hline
    $ asymmetric $ & &  &\\
    $ \tau_\alpha, \tau_\beta=$ & $\mathcal{L}_\texttt{symmetric}$ & 67.1 & 47.1 \\
     $(0.9, 0.6)$& $\mathcal{L}_\texttt{asymmetric}$ & \textbf{84.8} & \textbf{64.8}\\
    \bottomrule
    
    \end{tabular}
    \label{tab:ablation:loss}
  \end{minipage}
  \begin{minipage}[t]{0.55\linewidth}
   \centering
        \makeatletter\def\@captype{table}\makeatother\caption{Impact of the intra-sampling temporal interval and different training strategy. Evaluated on Market-1501.}
         \footnotesize
    \centering
    \begin{tabular}{l | l| c  c  c}
    \toprule
    Training &  & R-1 & mAP  \\
    \hline
    \multirow{4}{*}{Stage~I Only}& \scriptsize{\emph{Interval:}}& &  \\
    &\quad \ \  2 sec. & \textbf{29.4} & \textbf{11.9} \\
    &\quad \ \  4 sec. & 25.8 & 9.7 \\
     &\quad \ \  8 sec. & 27.2 & 10.4\\

     \hline
    \multirow{3}{*}{Stage~II Only}& \scriptsize{\emph{Data:}}& &  \\
     & intra + inter&  84.6 & 64.7\\
     &\quad \ \  inter & - & -  \\
     \hline
     \multirow{4}{*}{Stage~I + Stage~II}& \scriptsize{\emph{Interval:}}& &  \\
     &\quad \ \  2 sec. & \textbf{84.8} & \textbf{64.8}  \\
     &\quad \ \  4 sec. & 84.2 & 64.1 \\
     &\quad \ \  8 sec. & 84.0 & 53.3 \\
    
    \bottomrule
    
    \end{tabular}
      \label{tab:ablation:twostage}
   \end{minipage}
\end{minipage}
 \end{table}

    
\noindent \textbf{$\mathcal{L}_\texttt{symmetric}$ \vs $\mathcal{L}_\texttt{asymmetric}$. } To prove the proposed relaxed loss $\mathcal{L}_\texttt{asymmetric}$ is superior to the original loss $\mathcal{L}_\texttt{symmetric}$, we compare the two losses with  both symmetric and asymmetric training data. Results on Market dataset is shown 
in Table~\ref{tab:ablation:loss}.
We observe that with both symmetric/asymmetric training data, $\mathcal{L}_\texttt{asymmetric}$ consistently outperforms $\mathcal{L}_\texttt{symmetric}$. These results reveal a misunderstanding that $\mathcal{L}_\texttt{symmetric}$ is better for symmetric data and $\mathcal{L}_\texttt{asymmetric}$ is better for asymmetric data. 
In contrast, the results suggest $\mathcal{L}_\texttt{asymmetric}$ is essentially a more reasonable objective for the CycAs task, and it results in a more desirable embedding space. We speculate $\mathcal{L}_\texttt{symmetric}$ is too rigorous for the task, and the large magnitudes of losses from those well-associated cycles hamper the training.

\noindent \textbf{Impact of the temporal interval for intra-sampling. } We investigate the impact of the temporal interval for intra-sampling and show comparisons in Table~\ref{tab:ablation:twostage}. When only intra-sampled data are used for training (Stage I only), the sampling interval slightly affects the final performance. However, longer temporal interval, \eg, 8 seconds, does not bring performance gain.We speculate the reason could be the large asymmetry brought by long temporal interval .


    
    


\noindent \textbf{Two-stage training strategy. } The effectiveness of the proposed two-stage training strategy is also investigated, and results are shown in Tabel~\ref{tab:ablation:twostage}. For comparison, we first train with Stage I only, \ie, only intra-sampled data are used. It can be observed the final accuracy is quite poor. As discussed in previous section, the reason is  training with intra-sampled data does not align with the objective of the re-ID task, \ie, cross-camera retrieval. We also compare with training with Stage II only. If both intra- and inter-sampled data are used, the model converges to a decent solution. However, if we remove intra-sampled data from training, the model fails in converging. This suggest training with intra-sampled data is necessary for converging to a meaningful solution. Finally, if we use the proposed two-stage training, the results are as good as training with Stage II only, and the benefit is that the model converges with a faster speed.

\subsection{Experiment Using Self-collected Videos as Training Data}
\label{sec:exp:prac}
To our knowledge, prior works in unsupervised re-ID usually evaluate their systems on standard benchmarks by simulating real-world scenarios. 
In this paper, to further assess the practical potential of CycAs, we report experimental results obtained by training with real-world videos. 
The videos are captured by hand-hold cameras in several scenes with high pedestrian density, such as the airport, shopping mall and campus. The total length of the videos is about $6$ hours. Among these videos, about $5$ hours are captured from a single view, which can only be used for intra-sampling; The rest $1$ hour videos are captured from two different views, and thus can be used for inter-sampling. 

We employ an open-source pedestrian detector~\cite{jde} to detect persons in every $7$ frames and crop the detected persons. For intra-sampling, we set the maximum temporal interval between two frames to $2$ seconds. In every mini-batch, we sample $8$ frame pairs to enlarge the batch size for high training efficiency. 
Training lasts for $10$k iterations for Stage~I and another $35$k iterations for Stage~II. 

\begin{table}[t]
    \centering
    \begin{tabular}{l||c || p{1.2cm}<{\centering} |p{1.2cm}<{\centering} ||  p{1.2cm}<{\centering}  | p{1.2cm}<{\centering}    }
    \hline
           \multirow{2}{*}{} & \multirow{2}{*}{Training data}  &   \multicolumn{2}{c||}{Market-1501}  & \multicolumn{2}{c}{DukeMTMC}  \\
           \cline{3-6}
           &&  mAP  &R1 & mAP  & R1  \\
           \hline
           \hline
        
         Supervised.M & Market-1501~\cite{market} & 73.9 & 89.2 &  \emph{16.6} & \emph{33.4} \\
         Supervised.D & DukeMTMC~\cite{duke} & \emph{21.4} &  \emph{48.1} & 63.1 & 80.0 \\
         Supervised.C & CUHK03~\cite{cuhk03} & 19.8 &  43.2 & 13.1 & 26.3 \\
         \hline
         BUC~\cite{buc} & 6-hour unlabelled videos & {14.2} &  {29.8} & 11.2 & 21.5 \\
         UGA~\cite{uga} & 6-hour unlabelled videos & 17.8 &  37.2 & 15.4 & 25.6 \\
         \hline
         CycAs & 6-hour unlabelled videos & \textbf{23.3} &  \textbf{50.8} & \textbf{19.2} & \textbf{34.6} \\ 
         \hline
         
    \end{tabular}
    \caption{Results with real-world videos as training data. Comparisons are made with supervised baselines and existing unsupervised methods.}
    \label{tab:exp:real}
    
\end{table}

We evaluate the performance on Market-1501~\cite{market} and DukeMTMC-ReID~\cite{duke} test sets. Note that we do not use the training sets of Market-1501 and DukeMTMC-ReID. Since we use self-collected videos that are under completely different environments from Market-1501 and Duke-ReID, there is a large domain gap between our training data (self-collected video) and the test data. Results are presented in Table~\ref{tab:exp:real}. We make two observations. 

First, our method is significantly superior to unsupervised methods BUC~\cite{buc} and UGA~\cite{uga}. Both models are trained on our self-collected videos for fair comparison. For BUC we use the public code; For UGA we use our own implementation, and employ the JDE~\cite{jde} tracker to generate the tracklets. 
CycAs outperforms UGA and BUC by $+13.6\%$, and $+21.0\%$ in rank-1 accuracy on Market. It shows the promising potential of CycAs in real-world applications. 

Second, our method is very competitive or slightly superior to supervised models. For example, when trained on Market-1501 and tested on DukeMTMC, the IDE model obtains an mAP of 16.6\%. In comparison, we achieve 19.2\% mAP on DukeMTMC, which is +2.6\% higher. Similarly, our test performance on Market-1501 is $1.9\%$ higher than IDE trained on DukeMTMC. Moreover, our results on Market-1501 and DukeMTMC are consistently higher than IDE trained on CUHK03. In this experiment, the strength of CycAs lies in two aspects: 1) We utilize more training data. 2) We utilize unlabeled data in a more effective way. We also note that IDE is significantly higher than our method when IDE is trained and tested on the same domain. We reasonably think that if our system can be trained in a similar environment to Market-1501 or DukeMTMC (with properly deployed cameras), we would have a much better accuracy with a smaller domain gap. 



\section{Conclusion}
This paper presents CycAs, a self-supervised person ReID approach. 
We carefully design the pretext task---cycle association---to closely match the objective of re-ID. Objective function relaxations are made to allow end-to-end learning and introduce higher appearance variations in the training data. 
We train CycAs using public data and self-collected videos, and both settings validate the competitive performance of CycAs. 

\clearpage
%
%
\bibliographystyle{splncs04}
\bibliography{egbib}
\end{document}